\documentclass{article}
\usepackage{PRIMEarxiv}

\usepackage[utf8]{inputenc} %
\usepackage[T1]{fontenc}    %
\usepackage{hyperref}       %
\usepackage{url}            %
\usepackage{booktabs}       %
\usepackage{amsfonts}       %
\usepackage{nicefrac}       %
\usepackage{microtype}      %
\usepackage{lipsum}
\usepackage{fancyhdr}       %
\usepackage{graphicx}       %
\graphicspath{{media/}}     %

\usepackage[american]{babel}

\usepackage{natbib} %
\usepackage{mathtools} %
\usepackage{booktabs} %
\usepackage{amsfonts}       %
\usepackage{nicefrac}       %
\usepackage{microtype}      %
\usepackage{xcolor}         %
\usepackage{ bbold }
\usepackage[shortlabels]{enumitem}
\usepackage{ amssymb }
\usepackage{geometry}
\usepackage{bbm}
\usepackage{bm}

\usepackage{amsmath,amsthm}
\usepackage[utf8]{inputenc}
\usepackage{ulem}
\usepackage{tikz}
\usetikzlibrary{matrix, positioning}
\usetikzlibrary{automata,arrows}
\usepackage{listings}
\usepackage{mathtools}
\usepackage{verbatim}
\usepackage{subcaption}

\usepackage{algorithmicx}
\usepackage{algorithm}
\usepackage{algpseudocode}
\author{
  Bijan Mazaheri \thanks{Work done while interning at Amazon Research in Tübingen, Germany.}\\
  California Institute of Technology \\
  Pasadena, CA\\
  \texttt{bmazaher@caltech.edu} \\
   \And
  Atalanti Mastakouri, Dominik Janzing, Mila Hardt \\
  Amazon Research \\
  Tübingen, Germany\\
   \texttt{\{janzind, atalanti, milaha\}@amazon.de} \\
}

\title{Causal Information Splitting: \\ Engineering Proxy Features for Robustness to Distribution Shifts}

\let\R\Real

\def\argmin{\operatornamewithlimits{arg\,min}}

\newcommand{\indep}{\perp \!\!\! \perp}
\newcommand{\G}{\mathcal{G}}

\def\De{\operatorname{\mathbf{DE}}}

\def\An{\operatorname{\mathbf{AN}}}
\def\Ch{\operatorname{\mathbf{CH}}}

\def\Pa{\operatorname{\mathbf{PA}}}

\def\Fm{\operatorname{\mathbf{FM}}}

\def\pa{\operatorname{\mathbf{pa}}}

\def\Nb{\operatorname{\mathbf{NB}}}

\def\Pr{\operatorname{Pr}}
\def\I{\operatorname{\mathcal{I}}}
\def\H{\operatorname{\mathcal{H}}}

\makeatletter
\def\Bigbar#1{\mathrel{\left|\vphantom{#1}\right.\n@space}}

\makeatother

\newtheorem{theorem}{Theorem}
\newtheorem{lemma}{Lemma}

\theoremstyle{definition}
\newtheorem{definition}{Definition}

\renewcommand{\vec}{\bm}

\newcommand{\suppress}[1]{}

\def\given{\mid}

\usetikzlibrary{decorations,decorations.pathmorphing, decorations.pathreplacing, decorations.shapes, arrows, arrows.meta}

\pgfdeclaremetadecoration{middlezigzag}{straight}{
    \state{straight}[switch if less than=\pgfmetadecorationsegmentlength to final,
                  width=\pgfmetadecoratedpathlength/2 - \pgfmetadecorationsegmentlength/2 ,
                  next state=zigzag] {
        \decoration{curveto}
    }
    \state{zigzag}[width=\pgfmetadecorationsegmentlength,
                   next state=final] {
        \decoration{zigzag}
    }
    \state{final}{
        \decoration{curveto}
        \beforedecoration{\pgfpathmoveto{\pgfpointmetadecoratedpathfirst}}
    }
}

\tikzset{
    middle zigzag/.style={
        decorate,
        decoration={
            middlezigzag,
            meta-segment length=.36cm,
            segment length=0.12cm,
            amplitude = .05cm,
        }
    }
}

\newcommand\doubleactivepathNB{\mathrel{
\begin{tikzpicture}[baseline={([yshift=-.8ex]current bounding box.center)}]
    \draw [decorate, decoration={zigzag, segment length=.12cm, amplitude=.05cm}, double distance=.2mm] (-.25,0) -- (-.1,0);
    \draw [] (-.242,-.02) -- (-.218,0.02]);
    \draw [] (-.24,-.017) -- (-.25, -.017);
    \draw [-{Bar[width = 2mm]}, double distance=.2mm](-.1,0) -- (-.05,0);
    \draw [-{Circle[open, width=1.2mm, length=1.2mm]}, double distance=.2mm](-.25,0) -- (-.44,0);
\end{tikzpicture}}}
\newcommand\doubleactivepathBN{\mathrel{
\begin{tikzpicture}[baseline={([yshift=-.8ex]current bounding box.center)}]
    \draw [decorate, decoration={zigzag, segment length=.12cm, amplitude=.05cm}, double distance=.2mm] (.25,0) -- (.1,0);
    \draw [] (.242,.02) -- (.218,-0.02]);
    \draw [] (.24,.017) -- (.25, .017);
    \draw [-{Bar[width = 2mm]}, double distance=.2mm](.1,0) -- (.05,0);
    \draw [-{Circle[open, width=1.2mm, length=1.2mm]}, double distance=.2mm](.25,0) -- (.44,0);
\end{tikzpicture}}}

\newcommand\doubleactivepathNC{\mathrel{
\begin{tikzpicture}[baseline={([yshift=-.8ex]current bounding box.center)}]
    \draw [decorate, decoration={zigzag, segment length=.12cm, amplitude=.05cm}, double distance=.2mm] (-.25,0) -- (-.1,0);
    \draw [] (-.242,-.02) -- (-.218,0.02]);
    \draw [] (-.24,-.017) -- (-.25, -.017);
    \draw [-{to[width = 2mm]}](-.055,0) -- (-.05,0);
    \draw [-, double distance=.2mm](-.08,0) -- (-.1,0);
    \draw [-{Circle[open, width=1.2mm, length=1.2mm]}, double distance=.2mm](-.25,0) -- (-.44,0);
\end{tikzpicture}}}
\newcommand\doubleactivepathCN{\mathrel{
\begin{tikzpicture}[baseline={([yshift=-.8ex]current bounding box.center)}]
    \draw [decorate, decoration={zigzag, segment length=.12cm, amplitude=.05cm}, double distance=.2mm] (.25,0) -- (.1,0);
    \draw [] (.242,.02) -- (.218,-0.02]);
    \draw [] (.24,.017) -- (.25, .017);
    \draw [-{to[width = 2mm]}](.055,0) -- (.05,0);
    \draw [-, double distance=.2mm](.08,0) -- (.1,0);
    \draw [-{Circle[open, width=1.2mm, length=1.2mm]}, double distance=.2mm](.25,0) -- (.44,0);
\end{tikzpicture}}}

\makeatletter
\newcommand{\bigcomp}{%
  \DOTSB
  \mathop{\vphantom{\sum}\mathpalette\bigcomp@\relax}%
  \slimits@
}
\newcommand{\bigcomp@}[2]{%
  \begingroup\m@th
  \sbox\z@{$#1\sum$}%
  \setlength{\unitlength}{0.9\dimexpr\ht\z@+\dp\z@}%
  \vcenter{\hbox{%
    \begin{picture}(1,1)
    \bigcomp@linethickness{#1}
    \put(0.5,0.5){\circle{1}}
    \end{picture}%
  }}%
  \endgroup
}
\newcommand{\bigcomp@linethickness}[1]{%
  \linethickness{%
      \ifx#1\displaystyle 2\fontdimen8\textfont\else
      \ifx#1\textstyle 1.65\fontdimen8\textfont\else
      \ifx#1\scriptstyle 1.65\fontdimen8\scriptfont\else
      1.65\fontdimen8\scriptscriptfont\fi\fi\fi 3
  }%
}

\makeatother

\def\G{\mathcal{G}}

\def\Ugood{\vec{U}^{\mathrm{GOOD}}}

\def\Ubad{\vec{U}^{\mathrm{BAD}}}

\def\Vgood{\vec{V}^{\mathrm{GOOD}}}
\def\Vbad{\vec{V}^{\mathrm{BAD}}}
\def\Vambig{\vec{V}^{\mathrm{AMBIG}}}

\def\E{\mathbb{E}}
\def\Vsap{V_{\mathrm{SAP}}}
\newcommand{\Fisolate}[2]{F_{\mathrm{ISO}(#1)}(#2)}
\newcommand{\TFisolate}[2]{\tilde{F}_{\mathrm{ISO}(#1)}(#2)}

\usepackage{subfiles} %

\begin{document}
\maketitle
\begin{abstract}
Statistical prediction models are often trained on data from different probability distributions than their eventual use cases. One approach to proactively prepare for these shifts harnesses the intuition that causal mechanisms should remain invariant between environments. Here we focus on a challenging setting in which the causal and anticausal variables of the target are unobserved. Leaning on information theory, we develop feature selection and engineering techniques for the observed downstream variables that act as proxies. We identify proxies that help to build stable models and moreover utilize auxiliary training tasks to answer counterfactual questions that extract stability-enhancing information from proxies. We demonstrate the effectiveness of our techniques on synthetic and real data.
\end{abstract}

\section{Introduction}
The principle assumption when building any (not necessarily causal) prediction model is access to relevant data for the task at hand. When predicting label $Y$ from inputs $\vec{X}$, this assumption reads that the data is drawn from a (training) probability distribution $\vec{X}, Y$ 
that is identical to the distribution that will generate its use-cases (target distribution).

Unfortunately, the dynamic nature of real-world  systems
makes obtaining perfectly relevant data difficult. Data-gathering mechanisms can introduce sampling bias, yielding distorted training data. Even in the absence of sampling biases, populations, environments, and interventions give rise to distribution shifts in their own right. For example,~\cite{chest_xray_spurious} found that convolutional neural networks to detect pneumonia from chest radiographs often relied on site-specific features, including the metallic tokens indicating laterality and image processing techniques. This resulted in poor generalization across sites. Understanding these inter-site breakdowns in performance is essential to safety-critical domains such as healthcare.

\paragraph{Generalization and Invariant Sets}
The first attempts at handling dissociation between training and target distributions involved gathering unlabeled samples of the testing distribution. Within domain generalization (DG), covariate shift handles a shift in the distribution of $\vec{X}$ \citep{shimodaira2000improving} and label shift handles a shifting $\Pr(Y)$ \citep{schweikert2008empirical}. DG often assumes a stationary label function $\Pr(Y \given \vec{X})$, which is extremely limiting in real-life applications. To address these limitations, one can assume the label function is stationary for a \textit{subset} of the covariates in $\vec{X}$, called an \textbf{invariant set} in \cite{muandet2013domain} and \cite{rojas2018invariant}. 

One approach to finding invariant sets has been to capture shifting information from a collection of datasets \citep{rojas2018invariant, invariant_feature_selection}. Such techniques require access to a comprehensive set of datasets that represent all possible shiftings.
A causal perspective developed in \cite{storkey2009training} and \cite{pearl2011transportability} instead uses graphical modeling via \textbf{selection diagrams} to model shifting mechanisms. This approach requires access to multiple datasets to learn these mechanisms, but does not require that those datasets span the entire space of possible shifting. Such approaches also allow the use of domain expert knowledge when building selection diagrams. A detailed comparison of stability in the causal and anticausal scenario is given in \cite{scholkopf2012causal}.

\paragraph{Contributions}
The causal perspective to distribution shift is obscured when we lack direct measurements of the causes and effects of $Y$.  Such settings arise from noisy measurements, privacy concerns, as well as abstract concepts that cannot be easily quantified (such as ``work ethic'' or ``interests''). Instead, we will focus on a setting where we only measure \textit{proxies} for the causes and effects of $Y$, see Fig.~\ref{fig:example context} for an example. All of these proxies are descendants of $\vec{U}$ -- a case which is common in medicine, where the measured variables are often blood markers (or other tests) that are indicative of an underlying condition.

The proxy setting is difficult to address in standard framework. While previous approaches to partially observed systems suggest restricting model inputs to those on stable paths \citep{subbaswamy2018counterfactual}, no observed proxies satisfy this condition in our setting. That is, the proxies of the unobserved causes are insufficient to fully block the environmental shifts of those causes.

We will use concepts from causal inference and information theory to define and study the \textbf{Proxy-based Environmental Robustness (PER)} problem. Our framework will demonstrate that perfection is indeed the enemy of good -- some variables (although with an unstable relationship to the target) should still be included as features to build a model with improved stability.

A primary goal of this paper will be to distinguish between proxies that are ``helpful'' or ``hurtful'' for stability - a property that they inherit from their parents (of which they are proxies). The stability of these unobserved variables depends on their causal structure, which is unobserved. We will present a strategy for feature selection based on properties that propagate from the underlying causal structure to its observed proxies. 
Specifically, we will build on the observation that post-selecting on a single value of the prediction label $Y$ induces a special independence pattern, which the proxies also inherit. We use this to classify proxies from partial knowledge of a few ``seeds'' - a technique we call \textbf{proxy bootstrapping}.

It is possible that some proxy variables will contain information about both stable and unstable hidden variables. We call these \textbf{ambiguous proxies} because it is unclear whether they will improve or worsen the model's transportability. Inspired by node splitting~\citep{subbaswamy2018counterfactual}, we introduce a method we call \textbf{causal information splitting (CIS)}, which can improve stability of our models at no cost (and even some benefit) to the distribution shift robustness. CIS isolates stabilizing information using auxiliary prediction tasks that answer counterfactual questions about the covariates. While theoretical guarantees require a number of assumptions, we demonstrate the surprising ability of CIS to separate stabilizing information from ambiguous variables on synthetic data experiments with relaxed assumptions. Furthermore, we utilize CIS to enhance a prediction task on U.S. Census data that was strongly affected by the COVID-19 pandemic. While plenty of experiments have confirmed that techniques for robust models do not consistently provide benefits over empirical risk minimization~\citep{gulrajani2021in}, our proposed technique provides benefits for an income prediction task in the majority of tested states.
\section{Related Work}\label{sec:related work}
There is an increasing body of work on domain generalization, see~\cite{quinonero2008dataset} for an overview. 
While we focus on proactively modeling shifts, work on invariant risk minimization~\citep{arjovsky2019invariant,bellot2020generalization} has approached this problem when given access to the shifted data on which the models will be used. Recent work further generalizes to unseen environments constituting mixtures~\citep{sagawa2019distributionally} and affine combinations~\citep{krueger2021out}. Data from multiple environments can also be used for causal discovery~\citep{invariant_prediction, heinze2018invariant, peters2016causal}.

Another line of work seeks robustness to small adversarial changes in the input that should not change the output (with attacks, e.g.~\cite{Croce_adv_eval_attacks} and defenses, e.g.~\cite{sinha2018certifiable}). 
Moving from small changes to potentially bigger interventions, work on counterfactual robustness and invariance, introduces additional regularization terms~\citep{veitch2021counterfactual, CounterfactuallyInvPredictors}. 
 Our work differs by allowing for interventions that  change  the label.

We do not address the tradeoffs associated with robustness and model accuracy in this paper. Such tradeoffs are a natural consequence of restricting the input information for our model, since unstable information is still useful in unperturbed cases. This problem is generally addressed by \cite{oberst2021regularizing, rothenhausler2021anchor} via regularization.

The information theoretic decomposition presented in this paper is deeply related to the Partial Information Decomposition, which has been a subject of growing interest in information theory \citep{bertschinger2014quantifying, banerjee2018unique, williams2010nonnegative, gurushankar2022extracting, venkatesh2022partial}. The use of the PID in causal structures was pioneered by \cite{dutta2020information} for fairness \citep{dutta2023review}. Our method of extracting counterfactual features using auxiliary training tasks removes information from proxies that is unique to bad features, even though those features are not observed.

\section{Background}

\paragraph{General Notation}
Uppercase letters denote random variables, while lowercase letters denote assignments to those random variables. Bold letters denote sets/vectors. 
The paper will use concepts from information theory, with $\H(A)$ indicating the \textbf{entropy} of $A$, $\I(A : B)$ indicating the \textbf{mutual information} between $A, B$, and $\I(A:B:C)$ indicating the \textbf{interaction information} between $A, B, C$. A short summary of key ideas (including the data processing inequality (DPI) and chain rule) is given in Appendix B (see \cite{cover1999elements} for more details).

\paragraph{Causal Graphical Models}
Graphically modeling distribution shift makes use of causal DAGs. For a causal DAG $\G = (\vec{V}, \vec{E})$, the joint probability distribution factorizes according to the local Markov condition,
\begin{equation*}
 \Pr(\vec{v}) = \prod_{v \in \vec{v}} \Pr(v \given \pa_{\vec{v}}^\G(V)).
\end{equation*}

$\Pa^{\G}(V), \Ch^\G(V)$ denote the parents and children of $V$ in $\G$. Following the uppercase/lowercase convention, $\pa_{\vec{v}}(V)$ is an assignment to $\Pa(V)$ using the values in $\vec{v}$.\footnote{$\Pa(V) \subseteq \vec{V}$} $\De^{\G}(V)$ and $\An^{\G}(V)$ denote the descendants and ancestors respectively. $\Fm(V) = \Pa(V) \cup \Ch(V)$ denotes the ``family.''

We will rely on the concepts of $d$\textbf{-separation} and \textbf{active paths} to discuss the independence properties of Bayesian networks, which are discussed in Appendix A. See \cite{pearl2009causality} for a more extensive study.

\paragraph{Active Path Notation}
In addition to using $A \indep_d B \given C$ to indicate $d$-separation conditioned on $C$, we will develop a notation to refer to sets of variables that act as ``switches'' for $d$-separation. $\vec{A} \doubleactivepathNB \vec{C} \doubleactivepathBN \vec{B}$ means that we have both $\vec{A} \not \indep_d \vec{B}$ and $\vec{A} \indep_{d} \vec{B} \given \vec{C}$. Conversely, we have $\vec{A} \doubleactivepathNC \vec{C} \doubleactivepathCN \vec{B}$ if $\vec{A} \indep_d \vec{B}$, but $\vec{A} \not \indep_d \vec{B} \given \vec{C}$ (i.e. conditioning on $\vec{C}$ renders $\vec{A}$ and $\vec{B}$ $d$-connected).

\paragraph{Graphically Modeling Distribution Shift}
Borrowing terms from \cite{invariant_feature_selection}, we will begin with a graphical model $\G=(\vec{V} \cup \vec{U})$, calling $\vec{U} \cup \vec{V}$ the \textbf{system variables} with (un-)observed variables. 
In addition, we are also given a set of \text{context variables} $\vec{M}$, which model the mechanisms that shift our distribution. 
The augmentation of $\G$ with $\vec{M}$ gives what we call the \textbf{distribution shift diagram} (DSD), $\G^+ = (\vec{V} \cup \vec{U} \cup \vec{M}, \vec{E} \cup \vec{E}_{\vec{M}})$, for which $\G$ is a subgraph, with additional vertices $\vec{M}$ introducing shifts along $\vec{E}_{\vec{M}}$. $\vec{X} \subseteq \vec{V}$ such that $ \Pr(Y \given \vec{X}) = \Pr(Y \given \vec{X}, \vec{M})$ is called an ``invariant set'' because it blocks all possible influence from the mechanisms of the dataset shift. \cite{pearl2011transportability} shows this framework is capable of modeling sampling bias and population shift.
\section{Setting}
This paper will consider the \textbf{Proxy-based Environmental Robustness (PER)} setting. PER focuses on the role of proxy variables in feature selection by assuming \emph{all} of the causes and effects $\vec{U} = \Fm(Y)$ are unobserved.\footnote{This assumption is not necessary  but allows us to focus on more difficult questions that have not been answered by previous work. Namely, direct causes and effects can be visible or have perfect proxies without changing the results of the paper.} We are given access to a list of ``visible proxy variables'' $\vec{V} \setminus \{ Y \}$ which are descendants of at least one $U \in \vec{U}$.
Hence, $\vec{V}$ can be thought of as the union of overlapping subsets $\Ch(U)$ for each $U \in \vec{U}$. %

\begin{figure*}[tb]
 \centering
 (a)\scalebox{.5}{
 \begin {tikzpicture}[-latex ,auto ,node distance =2 cm and 2 cm ,on grid , thick, state/.style ={ circle, draw, minimum width =.85 cm}, cstate/.style ={ circle, draw, minimum width =.8 cm, ultra thick}]
  \node[state, accepting, color=green!50!black] (MU1) [] {$M_{1}$};
  \node[state, dashed, color=blue] (U1) [below right=of MU1] {$U_1$};
  \node[state] (V1)[below left = 3cm and 2cm of U1] {$V_1$};
  \node[state] (V2)[right = of V1] {$V_2$};
  \node[state] (V3)[ right = of V2] {$V_3$};
  \node[state] (V4)[ right = of V3] {$V_4$};
  \node[state] (V5)[ right = of V4] {$V_5$};
  \node[state] (V6)[ right = of V5] {$V_6$};
  \node[state] (V7)[ right = of V6] {$V_7$};
  \node[state, dashed, color=blue] (U2) [right = 4cm of U1] {$U_2$};
  \node[state, accepting, color=green!50!black] (MU2) [above left = of U2] {$M_{2}$};
  \node[state, dashed, color=blue] (U3) [right = 4cm of U2] {$U_3$};
  \node[state, accepting, color=green!50!black] (MU3) [above right = of U3] {$M_{3}$};
  \node[state] (Y) [ above = 4cm of U2] {$Y$};
  \path[very thick](U1) edge (V1) (U1) edge (V2) (U1) edge (V3) (U2) edge (V3) (U2) edge (V4) (U2) edge (V5) (U3) edge (V5) (U3) edge (V6) (U3) edge (V7);
  \path[very thick](U1) edge (V6) (U3) edge (V2) (U2) edge (V6);
  \path[very thick](U1) edge[bend left=30] (Y) (Y) edge (U2) (Y) edge[bend left=30] (U3);
  \path[very thick, color=green!50!black] (MU1) edge (U1) (MU2) edge (U2) (U3) edge (MU3);
 \end{tikzpicture}
 }
 (b)\scalebox{.67}{
 \begin {tikzpicture}[-latex ,auto ,node distance =2 cm and 2 cm ,on grid , thick, state/.style ={ rectangle, draw, minimum width =1 cm, thick, minimum height = .8cm}]
  \node[state] (Y) {Income};
  \node[state, green!50!black, double] (COVID) [below = 1.5cm of Y] {Pandemic};
  \node[state, dashed, blue] (Interests)[below left = 3cm and 4.5cm of Y] {Interests};
  \node[state, dashed, blue, minimum width =.8 cm] (Employment)[below left = 3cm and 1.5cm of Y]  {Employment};
  \node[state, dashed, blue, minimum width =.8 cm] (Residence)[below right = 3cm and 1.5cm of Y]  {Residence};
  \node[state, dashed, blue] (Eligibility)[below right = 3cm and 4.5cm of Y]  {Medicaid Eligibility};
  \node[state] (Commute)[below = 5cm of Y]  {Commute};
  \node[state] (MedicaidStatus)[below right = 5cm and 3cm of Y]  {Medicaid Status};
  \node[state] (Education)[below left = 5cm and 3cm of Y]  {Education};
  \path (Interests) edge[bend left = 20] (Y) (Employment) edge[bend left =20] (Y);
  \path (Y) edge[bend left = 20] (Residence) (Y) edge[bend left = 30] (Eligibility);
  \path (Interests) edge (Education);
  \path (Employment) edge (Commute) (Residence) edge (Commute);
  \path (Employment) edge (MedicaidStatus) (Eligibility) edge (MedicaidStatus);
  \path[green!50!black] (COVID) edge (Interests) (COVID) edge (Employment) (COVID) edge (Residence) (COVID) edge (Eligibility);
 \end{tikzpicture}}
 \caption{Examples of the $\G^+$ considered for the paper. (a) shows a generic setup where $U_1$ is a hidden cause of $Y$, and $U_2, U_3$ are hidden effects. (b) shows a \textit{plausible} model explaining the success of our real-data experiment in Section~\ref{sec:real_data}.}
 \label{fig:example context}
\end{figure*}
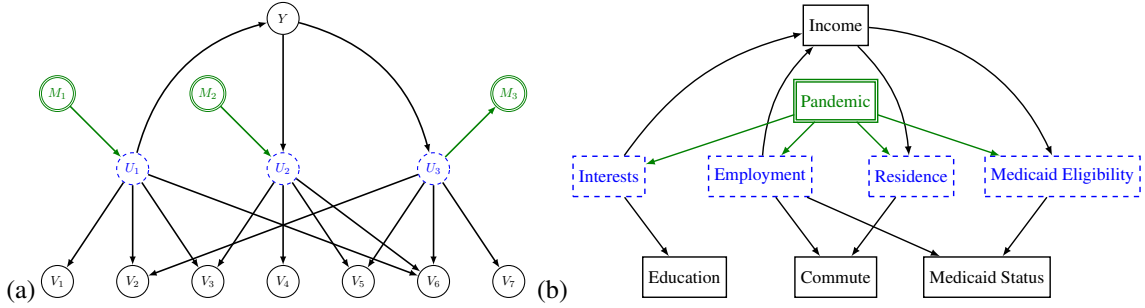

We will assume that there are no edges directly within $\vec{U}$ or within $\vec{V}$, which we call \textbf{systemic sparsity}. See Figure~\ref{fig:example context} for an example of this setting. This assumption enforces two useful independence properties: (1) $V_i \indep V_j \given U$ for $V_i, V_j \in \Ch(U)$ and (2) $U_i \indep U_j \given Y$ for $U_i \neq U_j \in \vec{U}$. Systemic sparsity guarantees that a discoverable causal structure exists within the unobserved variables and simplifies the interactions between the proxies.

We will build our theory on distribution shift diagrams $\G^+ = (\vec{V} \cup \vec{U} \cup \vec{M},  \vec{E} \cup \vec{E}_{\vec{M}})$  with one $M_i \in \vec{M}$ connected to a corresponding $U_i \in \vec{U}$. 
Each $M_i$ models a different shifting mechanism for each unobserved cause and effect of $Y$. It is common to assume there is no direct shifting mechanism acting on $Y$ - which comes without loss of generality since such a mechanism can be thought of as another unobserved cause \citep{pearl2011transportability, invariant_prediction}.

In this setting, a perfect invariant set $\vec{X}$ in which $Y \indep_d \vec{M} \given \vec{X}$ does not exist. PER will instead seek to minimize the influence of the context variables on our label function. Borrowing concepts from information theory, the task in PER corresponds to finding a set of features $\vec{X}$ that minimizes the conditional mutual information between the label and the environment. We call this quantity, $\I(Y:\vec{M} \given \vec{X})$, the \textbf{context sensitivity}. To allow for feature engineering, we define these features to be the output of a function, $\vec{X} = F(\vec{V} \setminus \{Y\})$ which can capture higher-level representations of $\vec{V} \setminus \{Y\}$.

\paragraph{Challenges in PER}
The PER setting is difficult to address using existing methods. Building a model on the causes $\Pa(Y)$ as in \cite{scholkopf2012causal} is impossible because all of the causes are unobserved. Furthermore, finding a separating set as in \cite{invariant_feature_selection, pearl2011transportability} is also impossible for the same reason.
Proxies can contain combinations of both stable and unstable information when they are connected to multiple $U \in \vec{U}$. Introduced in \cite{subbaswamy2018counterfactual}, ``node splitting'' requires knowledge of the structural equations that govern a vertex  to remove unstable information from ambiguous variables, which can only be learned if the causes of the split node are observed. This requirement limits node splitting's power in the proxy setting. 

\subsection{Invertible Dropout Functions}
We will demonstrate the failure of existing approaches in this setting using a counterexample built on structural equations models with cleanly interpretable entropic relationships. This construction will show the cost of restricting features to those with stable paths to the prediction variable $Y$, and serve as a framework for understanding the problem in general. For a discussion of relaxations, see Sec.~\ref{sec:conclusion} and for a demonstration that our method can work in real-world settings (where the assumption does not hold), see Sec.\ref{sec:aux}.

Our restricted structural equations give edges from $A$ to $B$ described by an invertible function with ``dropout'' noise,
\begin{equation} \label{eq: components structural eq}
    B^{(A)}(A) = \begin{cases}
        \mathcal{T}_{A,B}(A) &\text{with probability } \alpha_{A, B}\\
        \phi &\text{with probability } 1- \alpha
    \end{cases}.
\end{equation}
$\mathcal{T}_{A,B}(\cdot)$ is a function that is invertible, with $\mathcal{T}_{A,B}(\phi) = \phi$. The probability that information from the parent is preserved is given by $\alpha_{A,B} \in [0, 1]$. We will refer to $B^{(A)}(A) \neq \phi$ as ``transmission,'' and $\alpha_{A,B}$ as the ``probability of transmission.''\footnote{The direction of the edge for these $\alpha_{A,B}$ will sometimes be arbitrary, in which case the ordering of the vertices is unimportant.} $\phi$, called ``null'', is a value that represents the dropout, or the failure of the edge to ``transmit''. 

The structural equation for a vertex $B$ given its parents is a deterministic function of these $B^{(A)}$,
\begin{equation}
    B = \mathcal{T}_B(\{B^{(A)}(A) \text{ for } A \in \Pa(B)\}),
\end{equation}
where $\mathcal{T}_B$ is not necessarily an invertible function.

For functions with many children, the probability that at least one of their children transmits is
\begin{equation}
    \alpha_{A,\Ch(A)} := 1 - \prod_{B \in \Ch(A)}(1- \alpha_{A,B}).
\end{equation}

\begin{figure}[t]
 \centering
  \scalebox{.5}{
 \begin {tikzpicture}[-latex ,auto ,node distance =2 cm and 2 cm ,on grid , thick, state/.style ={ circle, draw, minimum width =.85 cm}, cstate/.style ={ circle, draw, minimum width =.8 cm, ultra thick}]
\filldraw[color=black, fill=black!5, very thick](-.8,-1) rectangle (2.8, -3)
            node[right] {$B$};
  \node[state] (A1) {$A_1$};
  \node[state] (A2)[right=of A1] {$A_2$};
  \node[state] (BA1)[below = of A1] {$B^{(A_1)}$};
  \node[state] (BA2)[below = of A2] {$B^{(A_2)}$};
  \path[very thick] (A1) edge (BA1) (A2) edge (BA2);
  \node[state] (A1a)[left = 6cm of A1] {$A_1$};
  \node[state] (A2a)[right=of A1a] {$A_2$};
  \node[state, fill=black!5] (B)[below right = 2cm and 1cm of A1a] {$B$};
  \path[very thick] (A1a) edge (B) (A2a) edge (B);
 \end{tikzpicture}
 }
 \caption{A diagram showing separability.} \label{fig:example splitting} 
\end{figure}
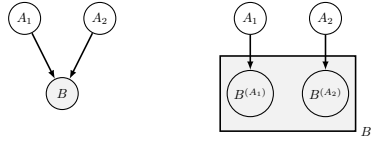

\paragraph{Separability and Faithfulness}
If $\mathcal{T}_B$ is invertible, we say that $B$ is a separable variable, which means that a child $B$ with more than one parent can be split into separate disconnected vertices $B^{(A)}$ for $A \in \Pa(B)$, each with the structural equation given by Equation~\ref{eq: components structural eq} (See Figure~\ref{fig:example splitting}). Separable variables make up a special violation of faithfulness in that conditioning on separable colliders no longer opens up active paths, illustrated by Lemma~\ref{lem: d conn but not dependent}.

\begin{lemma}[Separability violates faithfulness] \label{lem: d conn but not dependent}
If $U_1 \doubleactivepathNC V \doubleactivepathCN U_2$ and $V$ is separable, then $U_1 \not \indep_d U_2 \given V$, but $U_1 \indep U_2 \given V$.
\end{lemma}
The proof follows from the definition of mutual information and the fact that $U_1 \indep U_2 \given V$.

Our setting will rely on the assumption of faithfulness of the sub-graph on the $\vec{U} \cup \{Y\}$ vertices for proxy bootstrapping, as is the case for algorithms attempting any degree of structure learning. Specifically, we will require that any active path between two proxies $V_i, V_j$ that does not travel through any other vertices in $\vec{V}$ must imply statistical dependence (we call this ``partial faithfulness''). When we move to causal information splitting, we will allow \emph{specific} violations of faithfulness that come from separable proxies $\vec{V}$ in order to illustrate an ideal use-case of our method. This does not contradict partial faithfulness.

\paragraph{Transmitting Active Paths}
A convenient aspect of these structural equations is that $\alpha_{AB}$ controls the mutual information between $A$ and its child $B^{(A)}$,
\begin{equation*}
\begin{aligned}
    \I(A:B^{(A)}) =& \H(A) - \H(A \given B^{(A)})\\
    =& \H(A) - \Pr(B^{(A)} = \phi) \H(A \given B^{(A)}= \phi)\\
    &- \Pr(B^{(A)} \neq \phi) \H(A \given B^{(A)} \neq \phi)
\end{aligned}
\end{equation*}
An important insight is that $\H(A \given B^{(A)}= \phi) = 0$ and $\H(A \given B^{(A)} \neq \phi) = \H(A)$. Applying this gives,
\begin{equation*}
\begin{aligned}
    \I(A:B^{(A)}) = \H(A) - (1-\alpha_{A,B})\H(A) = \alpha_{A,B} \H(A).
\end{aligned}
\end{equation*}
This aspect generalizes to active single paths. For a length-2 path $A \rightarrow B \rightarrow C$, $
    \I(A:C) = \I(A:C^{(B)}) = \H(A) - \H(A \given C^{(B)})$.
Again, we can break up $\H(A \given C^{(B)})$ into $\H(A \given C^{(B)} = \phi) = 0$ and $\H(A \given C^{(B)} \neq \phi) = \H(A)$. Hence, reasoning about mutual information reduces to the task of determining the probability that one of the endpoints is null. In our setup, the dropout events of different edges are independent events. Hence, $\I(A:C) = \alpha_{A,B}\alpha_{B,C} \H(A)$.

Conditioning adds an additional complication. Notice that transmitting active paths can ``transfer'' a conditioning. That is, $\H(A \given x) = 0$ when there is only one active path between $A$ and $X$ (or $X$ to $A$) and it transmits. In the next section, we will study two cases that emerge in the PER problem: colliders and non-colliders.

\section{Context Sensitivity}
\label{sec:context_sensitivity}
We quantify robustness through the dependence on environmental mechanisms and the label function.

\begin{definition}[Context sensitivity]
Context sensitivity of a mechanism $M \in \vec{M}$  is defined as $\I(Y:M \given \vec{X})$. 
\end{definition}
If $\vec{X}$ $d$-separates $\vec{M}$ from $Y$, the context sensitivity is 0 and training on $\vec{X}$ to predict $Y$ yields a model that is robust across environments $\vec{M}$.

We are usually most concerned with the success of our prediction models, something that is limited by the ``relevance'', $\I(Y:\vec{X})$, of our input. This concept is related to context sensitivity, and we can rewrite the sensitivity in terms of the expected relevance across environments.
\begin{equation*}
\begin{split}
    \I(Y:M \given \vec{X}) = \I(Y:M) - \I(Y:M :\vec{X}) \\
    = \I(Y:M) - \I(Y:\vec{X}) + \I(Y:\vec{X} \given M) .
\end{split}
\end{equation*}

\subsection{Redundancy}
Recall that in our setting we assume that all direct causes and effects are unobserved. This unobserved set of parents
gives rise to an invariant set $\vec{S} \subseteq \vec{U}$\footnote{The Markov boundary of $Y$ would also give an invariant set, but could include vertices in $\vec{M}$ that are parents of effects of $Y$.}. We seek to identify a  subset of visible proxies $\vec{X} \subseteq \vec{V}$ to extract information about $\vec{S}$.

\begin{definition}
    For a specific $U$, we call $\I(U:\vec{X}) = \H(U) - \H(U \given \vec{X})$ the \textbf{redundancy} between $U$ and $\vec{X}$.
\end{definition}
\begin{lemma} \label{lem: setting redundancy}
In the dropout function setting, 
let $\Ch_{\vec{X}}(U) := \Ch(U) \cap \vec{X}$.
    \begin{equation*}
        \I(U:\vec{X}) = \alpha_{U,\Ch_{\vec{X}}(U)} \H(U).
    \end{equation*}
\end{lemma}
Redundancy in the dropout function setting is controlled by our choice of $\vec{X}$ via $\alpha_{U,\Ch_{\vec{X}}(U)}$, the probability of transmission to at least one child. 

Our graphical assumptions ensure that only one potential active path exists between each $M \in \vec{M}$ and $Y$ - hence each vertex acts as either a collider or a non-collider in the interaction of $M$ and $Y$ (and does not do both). We now demonstrate that redundancy with stable (non-collider) variables generally improves our context sensitivity, whereas redundancy with unstable (collider) variables worsens it. 
 
\paragraph{``Good'' $\vec{U}$} If $M_i$ and $Y$ do not form a collider at $U_i \in \vec{U}$, we say $U_i \in \Ugood$. From $d$-separation, we have that $M_i \indep_d Y \given U_i$ for all $U_i \in \Ugood$. For an example, $\Ugood = \{U_1, U_3\}$ in Figure~\ref{fig:example context}. Let $\Ch_{\vec{X}}(U_i) = \Ch(U_i) \cap \vec{X}$.

\begin{lemma}[Redundancy with $\Ugood$] \label{lem: new applied DPI}
In the dropout function setting, for some $U_i \in \vec{U}$, if corresponding $M_i \doubleactivepathNB U_i \doubleactivepathBN Y$, then
\begin{equation*}
    \I(M_i:Y \given \vec{X}) = \alpha_{M_i,U_i}(1 - \alpha_{U_i, \Ch_{\vec{X}}(U_i)})\alpha_{U_i,Y} \H(M_i).
\end{equation*}
\end{lemma}
Lemma~\ref{lem: new applied DPI} comes from multiplying the probability of transmission of each edge along the path $M_i, U_i, Y$. We also pick up a term requiring that the $U_i, \vec{X}$ edges do not transmit, in which case conditioning on $\vec{X}$ would reduce the entropy of $U$ to nothing and close off the path.

\paragraph{``Bad'' $\vec{U}$} The inclusion of $\Ch(U_i)$ in $\vec{X}$ could open up active paths via colliders of the form $M_i \rightarrow U_i \leftarrow Y$. We call the set of these variables $\Ubad$. For an example, $\Ubad = \{U_2\}$ in Figure~\ref{fig:example context}.
\begin{lemma}[Redundancy with $\Ubad$]
\label{lem: new Collider DPI}
In the dropout function setting, $U_i \in \vec{U}, \vec{X} \subseteq \vec{V}$, if $M_i \doubleactivepathNC U_i \doubleactivepathCN Y$ then 
\begin{equation*}
     \I(M_i:Y \given \vec{X}) = \alpha_{U_i, \Ch_X(U_i)} \I(M_i:Y \given U_i) 
\end{equation*}
\end{lemma}
Lemma~\ref{lem: new Collider DPI} demonstrates that there are still proxies for which inclusion hurts our model's robustness. Similar concepts can be demonstrated via upper bounds when we allow arbitrary sets of structural equations - given in Appendix C. Optimizing these upper bounds does not give a guarantee of optimality, but can still point towards a general improvement.

\subsection{Feature Selection Implications}
The proxy graphical setup requires $\vec{X} \doubleactivepathNB \vec{U} \doubleactivepathBN Y$, meaning the relevance of our input is upper bounded by the redundancy with $U$,
$\I(\vec{X}: Y) \leq \I(\vec{U} : \vec{X})$.

Lemma~\ref{lem: new applied DPI} shows that proxies of $\Ugood$ help build accurate and universal models, while Lemma~\ref{lem: new Collider DPI} shows that proxies of $\Ubad$ can trade universality for domain-specific accuracy. Of course, proxies need not lie neatly in these two classes - many proxies contain a combination of universally-relevant and domain-relevant features. This suggests multiple classes of proxy variables.
\begin{definition}
    \begin{align}
        \Vgood &:= \Ch(\Ugood) \setminus \Ch(\Ubad)\\
        \Vbad &:= \Ch(\Ubad) \setminus \Ch(\Ugood)\\
        \Vambig &:= \Ch(\Ubad) \cap \Ch(\Ugood)
    \end{align}
\end{definition}

The behavior of $\Vgood$ in the dropout function setting shows how restricting models to invariant features fails; a high redundancy with $\Ugood$ is beneficial for the context sensitivity even though the paths from the proxies are unstable. Inclusion of $\Vgood$ in $\vec{X}$ improves context sensitivity even though $\Vgood$ is not made up of direct causes (as suggested by \cite{scholkopf2012causal}) or invariant features (as suggested by \cite{invariant_feature_selection} and \citep{subbaswamy2018counterfactual}).

For feature selection, an obvious strategy is to choose $\vec{X} = \Vgood$, avoid $\Vbad$, and potentially try using some elements in $\Vambig$. In the next section we will explore how we can use non-invertible functions to transform these $\Vambig$ into $\Vgood$.

\subsection{Proxy Bootstrapping}\label{sec:bootstrapping}
Given the robustness implications of the different classes of $V$, their partitioning into good, bad, and ambiguous partitions will be important. We will now demonstrate how to harness partial information to determine these partitions and classify proxies. This step is optional if the role of each proxy is already understood (as is the case when the DAG is known). The results in this subsection will \emph{only} require the graphical assumptions of the PER setting - i.e. systemic sparsity, partial faithfulness, and an independent shifting mechanism $M_i$ for each $U_i \in \vec U$. 

We begin with an observation about the independence structure of the conditional probability distribution on $Y$.
\begin{lemma}[Linking related proxies]\label{lem:conditioning on y separates sets}
Within the graphical constraints of PER, if $V_i \not \indep_d V_j \given Y$, then either they have a shared parent ($\Pa(V_i) \cap \Pa(V_j) \neq \emptyset$) or they both have at least one parent that is a cause of $Y$ (i.e. $\Pa(V_i) \cap \Pa(Y) \neq \emptyset$ and $\Pa(V_j) \cap \Pa(Y) \neq \emptyset$).
\end{lemma}

\begin{definition}\label{def: DependenceGraph}
    For a DSD $\G^+ = \{\vec{V} \cup \vec{U}\cup \vec{M}, \vec{E}\}$, define the dependence graph $\G_Y = (\vec{V}, \vec{E}_Y)$ to be an undirected graph with edges $(V_i, V_j) \in \vec{E}_Y$ iff $V_i \not \indep_d V_j \given Y$.
\end{definition}

Lemma~\ref{lem:conditioning on y separates sets} tells us that $\G_Y$ will have a clique on the sets $\Ch^\G(U)$ for $U \in \vec{U}$. Furthermore, conditioning on $Y$ links its causes, so $\G_Y$ has one large clique on $\Ch^\G(\Pa(Y))$. This clique structure can be utilized to enhance partial knowledge of $\Ch(\Ugood)$ and $\Ch(\Ubad)$. In this sense, ``birds of a feather flock together'' -- information about each clique's proxies can be a determined from understanding a single member of that clique.
 \begin{lemma}[Information about seed proxies spreads]\label{lem: bootstraping} If $V_i \in \Vgood$ then all neighbors of $V_j \in \Nb^{\G_Y}(V_i)$ are not in $\Vbad$ - i.e. $V_j \in \Vgood \cap \Vambig$. If $V_i \in \Vbad$ then all neighbors of $V_j \in \Nb^{\G_Y}(V_i)$ are not in $\Vgood$ - i.e. $V_j \in \Vbad \cap \Vambig$.
 \end{lemma}
 
 Lemma~\ref{lem: bootstraping} suggests a simple algorithm for bootstrapping the sets $\Vgood, \Vbad, \Vambig$ from a set of ``seed'' vertices $\vec{V}^* \subseteq \vec{V}$ with known assignments to $\Vgood, \Vbad, \Vambig$.

\begin{enumerate}
    \setlength{\itemsep}{0pt}
     \item Construct $\G_Y$ according to Definition~\ref{def: DependenceGraph} using conditional independence tests.
     \item For each $V^* \in \vec{V}^*$, if $V^* \in \Vgood$ then add a ``good'' label to $\Nb(V^*)$. If $V^* \in \Vbad$ then add a ``bad'' label to $\Nb(V^*)$.
     \item All $V \in \vec{V} \setminus \vec{V}^*$ with both ``good'' and ``bad'' labels receive an ``ambigious'' label instead.
\end{enumerate}
 
 \begin{theorem}[Proxy bootstrapping works]\label{thm: proxy bootstrapping works}
 Upon termination of proxy bootstrapping all vertices with a single label are correctly described if :
 \begin{enumerate}
    \setlength{\itemsep}{0pt}
    \item Partial faithfulness holds.
     \item $\vec{V}^*$ has at least one $V^* \in \vec{V}^* \cap \Ch(U)$ for each $U \in \Ugood \cap \Ch(Y)$.
     \item $\vec{V}^*$ has at least one $V^* \in \vec{V}^* \cap \Ch(\Pa(Y))$.
     \item $\vec{V}^*$ has at least one $V^* \in \vec{V}^*$ for each $U \in \Ubad$.
 \end{enumerate}
 \end{theorem}
\renewcommand{\Vsap}{V_{A}}

\section{Causal Information Splitting}
This section will expand our theory into \textbf{feature engineering}, which allows us to build inputs on functions of $\vec{V}$. A main takeaway from Section~\ref{sec:context_sensitivity} was that we should build models using proxies for $\Ugood$ and avoid using features that are proxies for $\Ubad$. The extension of this to engineered features is to build a model on \emph{functions} of proxies for which the output of those functions \emph{is} related to $\Ugood$ and \emph{not} related to $\Ubad$. We present two lemmas to formalize this notion.

Let $\widetilde{\Ch}_{\vec{X}}(U_i)$ be the children or functions of children of $U_i$ in $\vec{X}$. Lemma~\ref{lem: new applied DPI engineering} shows that building models with more redundancy with $\Ugood$ (i.e. lower $\H(U_i \given \widetilde{\Ch}_{\vec{X}}(U_i)$) improves our context sensitivity in the dropout function setting.\footnote{Appendix C shows that redundancy with $\Ugood$ lowers an upper bound on context sensitivity in more general cases}

\begin{lemma}[Engineering redundancy for $\Ugood$] \label{lem: new applied DPI engineering}
In the dropout function setting, if $U_i \in \Ugood$ then
\begin{equation*}
    \I(M_i:Y \given \vec{X}) = \alpha_{M_i,U_i}\alpha_{U_i,Y}\H(U_i \given \widetilde{\Ch}_{\vec{X}}(U_i)).
\end{equation*}
\end{lemma}

Of course, even good proxies are related to $\Ubad$ through their connection to $Y$, so $\vec{X} \indep \Ubad$ is impossible. Instead, Lemma~\ref{lem: if we dont include information with bad u, then we are happy} tells us that if we avoid redundancy with $\Ubad$ after conditioning on $Y$, we do not pick up any context sensitivity from the associated shifting mechanisms.

\begin{lemma}[Avoiding redundancy with $\Ubad$] \label{lem: if we dont include information with bad u, then we are happy}
For some $U_i \in \Ubad$, if we maintain $\I(U_i: \vec{X} \given Y) = 0$, then $\I(M_i:Y \given \vec{X}) = 0$.
\end{lemma}

Recall that ambiguous proxies contain information about both $\Ugood$ and $\Ubad$. The inclusion of an ambiguous proxy $V_A$ improves context sensitivity because of its redundancy with $\Ugood$ via Lemma~\ref{lem: new applied DPI engineering}. This section will develop a technique for filtering $V_A$ into $F(V_A)$, which will satisfy the conditions in Lemma~\ref{lem: if we dont include information with bad u, then we are happy}. To do this, we will require separability.

\paragraph{Separable Ambigious Proxies}
\begin{figure}[h]
    \centering
    \scalebox{.45}{
    \begin {tikzpicture}[-latex ,auto ,node distance =2 cm and 1.5 cm ,on grid , thick, state/.style ={ circle, draw, minimum width =.85 cm}, cstate/.style ={ circle, draw, minimum width =.8 cm, ultra thick}]
            \filldraw[color=black, fill=black!5, very thick](-1.4,-3.6) rectangle (1.4, -5.4)
            node[right] {$V_A$};
            \node[state] (Y) [] {$Y$};
            \node[state, dashed, color=blue] (UG) [below left = 3cm and 2 cm of Y] {$U_G$};
            \node[state, dashed, color=blue] (UB) [below right=3cm and 2cm of Y] {$U_B$};
            \node[state, accepting, color=green!50!black] (MUG) [above left = 1.5cm and 2cm of UG] {$M_{G}$};
            \node[state, accepting, color=green!50!black] (MUB) [above left =1.5cm and 2cm of UB] {$M_{B}$};
            \node[state] (VG) [below right = 1.5cm and 1.35cm of UG] {$V_A^{(G)}$};
            \node[state] (V1) [below left = 1.5cm and 2cm of UG] {$V_G$};
            \node[state] (V2) [below right = 1.5cm and 2cm of UB] {$V_B$};
            \node[state] (VB) [below left = 1.5cm and 1.35cm of UB] {$V_A^{(B)}$};
            \path[very thick] (UG) edge (VG) (UG) edge (Y) (Y) edge (UB) (UB) edge (VB);
            \path[very thick] (MUG) edge (UG) (MUB) edge (UB);
            \path[very thick] (UG) edge (V1) (UB) edge (V2);
    \end{tikzpicture}
    }
    \caption{$V_G \in \Vgood$, $V_B \in \Vbad$. $V_A \in \Vambig$ is a linear transformation of two components, $V_A^{(G)},V_A^{(B)}$, which are good and bad respectively.\label{fig:ambiguous and splitable}} 
\end{figure}
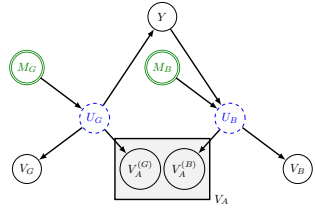
Consider the setup in Figure~\ref{fig:ambiguous and splitable}, where $V_G \in \Vgood$, $V_B \in \Vbad$, and $V_A \in \Vambig$. $V_A$ is generated by invertible $\mathcal{T}_A$, making it a \textbf{separable ambiguous proxy (SAP)}.\footnote{While we may still be able to gain useful information from non-separable proxies, the tradeoffs are difficult to quantify and hence beyond the scope of this paper.}
Splitting $V_A$ into components allows us to isolate the origins of its ambiguity - the mixing of good information from $V_A^{(G)}$ and bad information from $V_A^{(B)}$.

\subsection{Isolation Functions}\label{ssec:splitting}
We would like to isolate $V_A^{(G)}$ from $V_A$ to avoid paying the penalty for $V_A^{(B)}$. We will do this using \textbf{isolation functions}.

\begin{definition}%
We define an \textbf{isolation function} of $V_i$ on $V_A$, with optional conditioning on $y$, to be
\begin{equation}
    \begin{split}
    &\Fisolate{V_i}{V_A \given y}:=\argmin_F \H(F(V_A \given y))\\
    &\text{such that}  \I(F(V_A): V_i \given y) = \I(V_A : V_i \given y).
\end{split}
\end{equation}
$\Fisolate{V_i}{V_A \given Y}$ gives a vector of functions with an entry for each $y \in Y$.
\end{definition}
Note that isolation functions are sufficient statistics for $V_i$ \citep{cover1999elements}. Isolation involves maintaining the information about $V_i$ while removing excess noise.

Recall from Lemma~\ref{lem: if we dont include information with bad u, then we are happy} that in order to avoid worsening context sensitivity, we want to ensure $\I(F(V_A): \Ubad \given Y) = 0$. Isolation functions on SAPs are well designed for this purpose, because they enforce the independence properties of the isolated vertex on their outputs. In order to achieve $\I(F(V_A): \Ubad \given Y) = 0$ while preserving as much information about $\Ugood$ as possible, an optimal isolation function would be to isolate $\Ugood$ using $\Fisolate{\Ugood}{V_A \given Y}$. 

Of course, we do not have access to $\Ugood$, so our next best option is to isolate $\Vgood$ using $\Fisolate{\Vgood}{V_A \given Y}$, since $\Ubad \indep \Vgood \given Y$. Lemma~\ref{lem: isolation bound 2} shows that the output of $\Fisolate{V_G}{V_A \given Y}$ behaves like a good proxy if $V_G \in \Vgood$ and $V_A$ is a SAP.
\begin{lemma}[Isolating $\Vgood$ behaves like $\Vgood$]\label{lem: isolation bound 2}
For $V_G \in \Vgood$ and $U_B \in \Ubad$ and an isolation function $\Fisolate{V_G}{V_A \given Y}$, 
\[
   \I(U_B : \Fisolate{V_G}{V_A \given Y} \given Y) = 0.
\]
\end{lemma}
The benefit from $\Fisolate{V_G}{V_A\given Y}$'s information about $\Ugood$ is difficult to quantify for use with Lemma~\ref{lem: new applied DPI engineering}, but lower bounds are obtained in Appendix D.

Even without a quantification of improvement, Theorem~\ref{thm: when help and do no harm} shows that isolation functions can avoid worsening the context sensitivity, while certain conditions can guarantee relevance gains for predicting $Y$.
\begin{theorem}[CIS costs and benefits]\label{thm: when help and do no harm}
Consider $V_G \in \Vgood$ and $V_A \in \Vambig$ where $V_A$ is a SAP. Also consider the isolation function $\Fisolate{V_G}{V_A \given Y}$. We will compare the context sensitivity of inputs $\vec{X} := \{V_G\}$ and $\vec{X}^+ := \{V_G, \Fisolate{V_G}{V_A \given Y})\}$. We claim that $\I(M:Y \given \vec{X}^+) \leq \I(M:Y \given \vec{X})$ for all $M \in \vec{M}$.
Furthermore, if 
\begin{equation}\label{eq: thm2}
\begin{aligned}
    \I(\Fisolate{V_G}{V_A \given Y}:V_G) &<\\ \I(\Fisolate{V_G}{V_A \given Y}&:V_G \given Y),
\end{aligned}
\end{equation} then the relevance improves: $\I(Y:\vec{X}^+) > \I(Y:\vec{X})$.
\end{theorem}
Theorem~\ref{thm: when help and do no harm} tells us that using an isolation function helps when the function is more predictive of the isolated variable in the post-selected $Y$ distribution than it is in the full distribution. This condition is sufficient but loose  because it does not take into account direct effects from $\I(Y:\Fisolate{V_G}{V_A \given Y})$ (for which we have no guaranteed bounds). The proof is given in Appendix E.

\subsection{Auxiliary Training Tasks} \label{sec:auxtask}
In the infinite sample regime, consider an ``optimal'' model $F(\cdot)$ that predicts $V_i$ using input $V_A$. Optimal models should utilize all of the information available for prediction in their inputs, meaning $I(F(V_A): V_i) = \I(V_A: V_i)$. Information theoretically, minimizing $\H(F_{V_i}(V_A))$ corresponds to reducing the outputs of $F_{V_i}(V_i)$ to equivalence classes wherein $\Pr(V_A \given F_{V_i}(V_i) = f)$ is constant. This minimization corresponds to ensuring $F_{V_i}(V_i)$ does not over-fit to the empirical values of $V_A$ using noise that is orthogonal to $\Pa(V_A)$.

Auxiliary training tasks can therefore be used in place of isolation functions: we can get an approximate isolation function, $\TFisolate{V_i}{\Vsap}$, by training a model to predict $V_i$ using input $\Vsap$. We do not give any theoretical results beyond intuition for this interpretation, but will support our claims with experiments in the next section.

Equation~\ref{eq: thm2} in Theorem~\ref{thm: when help and do no harm} also has a nice interpretation within the training context -- the accuracy of the predictor must degrade when moving from the post-selected data to the full dataset. More precisely, the conditions for improvement now translate to
\begin{equation}
\begin{split}
    \min_F &\E[\mathrm{Error}(F(V_A), V_G)] \\
    &> \sum_{y} \Pr(y) \min_F(\E[\mathrm{Error}(F(V_A), V_G) \given y]),
\end{split}
\end{equation}
which can easily be checked on our training data.

\subsection{Suggested Overall Procedure}
We propose the following procedure for building robust (low context-sensitivity) models in the PER problem.
\begin{enumerate}
    \setlength{\itemsep}{0pt}
    \item Partition the data into constant $Y=y$ and determine cliques of dependence.
    \item Using domain knowledge, identify seeds in $\Vgood, \Vbad$ for proxy bootstrapping (Sec.~\ref{sec:bootstrapping}).
    \item Perform CIS on $\Vambig$  (Sec.~\ref{sec:auxtask}).
    \item Build a prediction model for $Y$ using $\Vgood$ and the CIS-engineered $\Vambig$.
\end{enumerate}

\section{Experiments} \label{sec:aux}
 We will now demonstrate the effectiveness of these methods on synthetic and real world data. Full code for both of these experiments is available at \url{https://zenodo.org/badge/latestdoi/651823136}.

\subsection{Experiments on Synthetic Data}
We generate data for the DAG in Figure~\ref{fig:ambiguous and splitable} based on normal distributions, see details of the setup in Appendix F. We vary the standard deviations of normally distributed $M_G$ and $M_B$.
The training data is drawn from $\sigma(M_G) = \sigma(M_B) = 1$, while the testing data varies both quantities and thus the influence of the context. We measure the accuracy of our feature engineering based on CIS, $\hat{Y}^{(3)}(V_G, \TFisolate{V_G}{V_A})$, that  utilizes the auxiliary task approximation to isolate $V_A$'s predictive information about $V_G$. We compare it to $\hat{Y}^{(1)}(V_G, V_A)$ trained on $\Vgood \cup \Vambig$ and $\hat{Y}^{(2)}(V_G)$ trained on only $\Vgood$. For a theoretical limit of CIS we also compare to  $\hat{Y}^{(4)}(V_G, V_A^{(G)})$ although access to $V_A^{(G)}$ is usually not possible.

\begin{figure}[ht]
    \centering
    (a)\includegraphics[width = .4\textwidth]{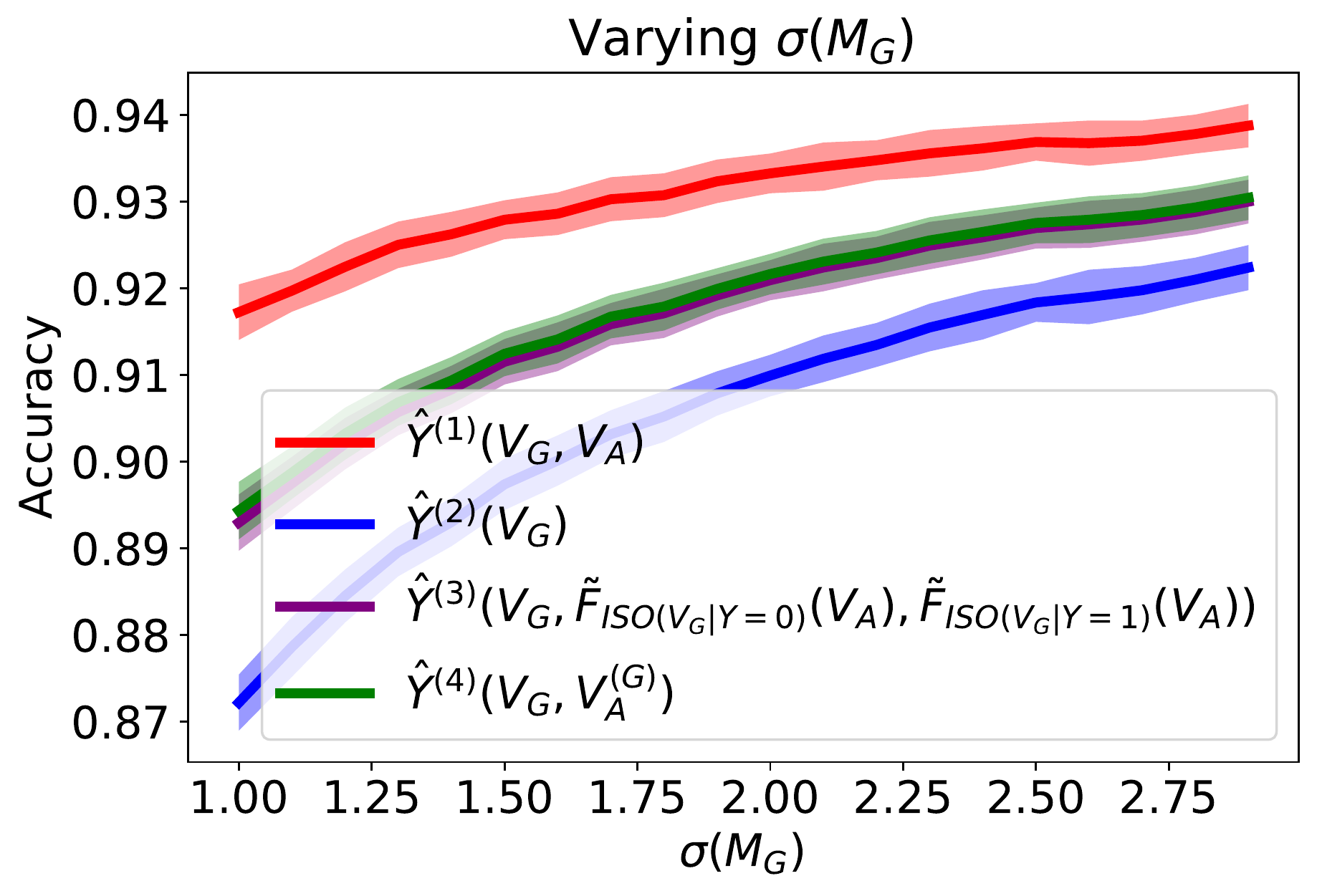}
    (b)\includegraphics[width = .4\textwidth]{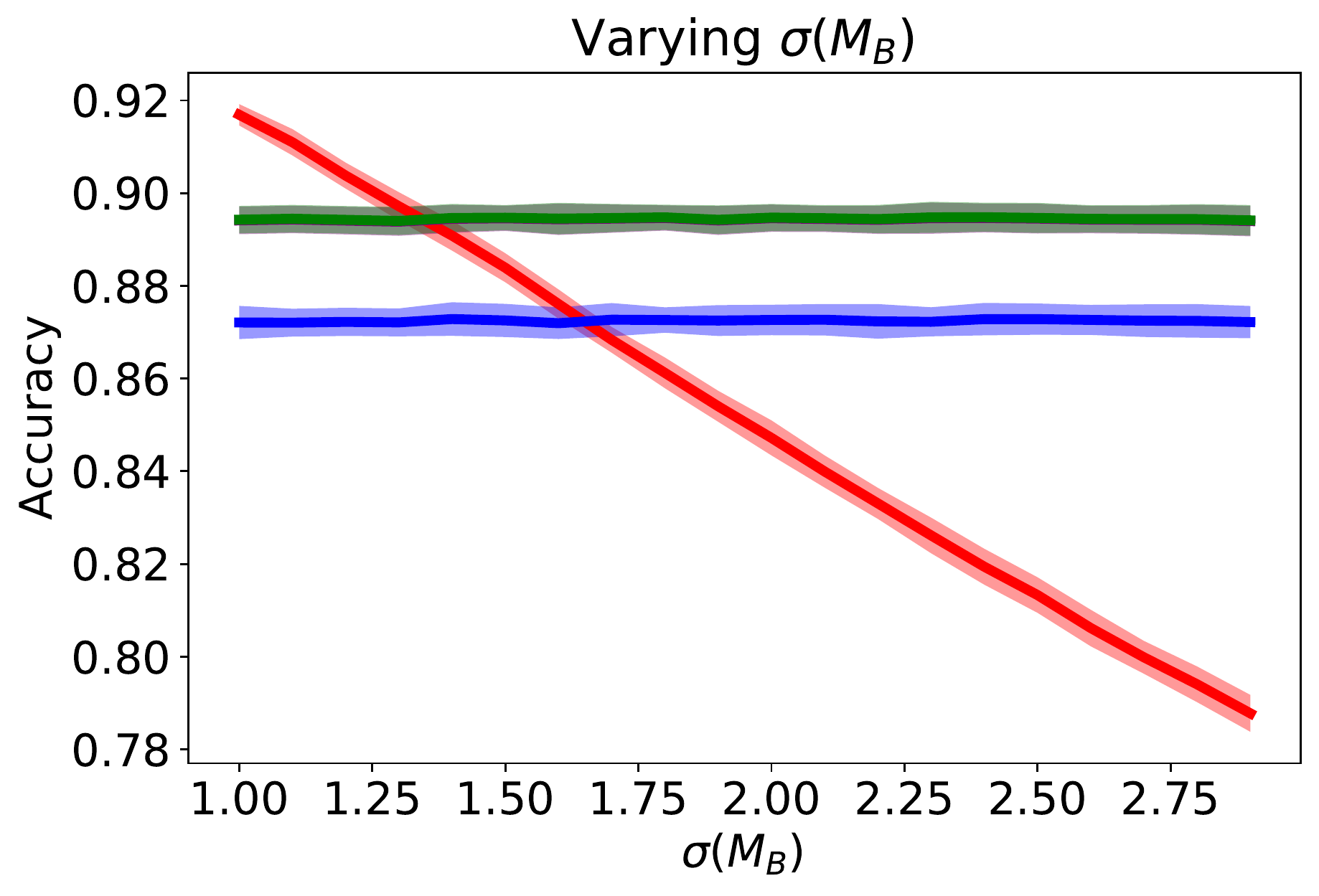}
    \caption{Results from our experiments on synthetic data. Single standard deviation confidence intervals are shaded in the corresponding colors.
    }
    \label{fig:experiments_proxies}
\end{figure}

\paragraph{Results} When comparing feature selection approaches, we observe in Figure~\ref{fig:experiments_proxies} that including $V_A$ results in higher accuracy of $\hat{Y}^{(1)}$ over $\hat{Y}^{(2)}$ when the shift acts on $\Ugood$ (a) or is small for $\Ubad$ (b). However, the accuracy of $\hat{Y}^{(2)}$ deteriorates with bigger shifts in $\Ubad$.

Our proposed method based on causal information splitting offers a middle ground.  $\hat{Y}^{(3)}$ is able to maintain the same robustness as $\hat{Y}^{(2)}$ while taking advantage of some of the gains enjoyed by $\hat{Y}^{(1)}$ in (a). In fact, $\hat{Y}^{(3)}$ performs very similarly to $\hat{Y}^{(4)}$, which had a-priori knowledge of the SAP components and used only $V_A^{(G)}$. These improvements were achieved despite not meeting the sufficient condition for increasing relevance in Theorem~\ref{thm: when help and do no harm}.

\subsection{Experiments on Census Data} \label{sec:real_data}
We use US Census data processed through folktables~\cite{ding2021retiring} to predict whether the income of a person exceeds 50k following ~\cite{Dua:2019}. %
To test out-of-domain generalization, prediction models were built on 2019 pre-pandemic data and evaluated on 2021 data during the pandemic.\footnote{We ignored the experimental release of 2020 data to ensure a starker distribution shift.} 
As model inputs, we consider commute time (coded as JWMNP in the dataset), a flag whether the person received Medicaid, Medical Assistance, or any kind of government-assistance plan for those with low incomes or a disability (coded as HINS4) and education level (SCHL). 
This small feature set was purposefully selected to see a starker effect of including/excluding individual features, including  a feature with relatively stable predictive power (education level) and two features heavily affected by the pandemic through increased work-from-home and medicaid's continuous enrollment provision.  

Our auxiliary task from Sec.~\ref{sec:auxtask}, referred to as \emph{engineered features}, does not use HINS4 and JWMNP directly as input features to predict the income level. Instead it uses HINS4 and JWMNP to train two models predicting the education-level: One trained on examples with high income and one trained on examples with low income. These predictions based on HINS4 and JWMNP together with the  actual education-level serve as input features to the final model.  
We compare the model built on these engineered features to ones using all three features directly (\emph{all features}) or using just the stable education feature (\emph{limited features}).
 
We use logistic regression from sklearn with l1 regularization to build models based on the different feature sets that the three methods created. l1 regularization yielded better generalization than l2 regularization.

\begin{table}[h!]
    \centering
    \small
    \caption{Comparison of out-of-domain (2021) performance via mean of accuracy.}
    \scalebox{.9}{
    \begin{tabular}{|c | l l l |  }
\hline
                         State & All Features & Engineered Features & Limited Features   \\ 
                        \hline
                    CA  & \textbf{0.712} $\pm$ 0.0011 & \textbf{0.711} $\pm$ 0.0014 & 0.692 $\pm$ 0.0014 \\
                    FL  & \textbf{0.683} $\pm$ 0.0012 & 0.678 $\pm$ 0.0018 & 0.680 $\pm$ 0.0013 \\
                    GA  & 0.689 $\pm$ 0.0025 & \textbf{0.707} $\pm$ 0.0055 & \textbf{0.709} $\pm$ 0.0029 \\
                    IL  & 0.662 $\pm$ 0.0026 & \textbf{0.689} $\pm$ 0.0033 & 0.684 $\pm$ 0.0019 \\
                    NY  & \textbf{0.707} $\pm$ 0.0022 & \textbf{0.702} $\pm$ 0.0025 & 0.687 $\pm$ 0.0080 \\
                    NC  & \textbf{0.691} $\pm$ 0.0031 & \textbf{0.684} $\pm$ 0.0034 & \textbf{0.683} $\pm$ 0.0030 \\
                    OH  & 0.689 $\pm$ 0.0022 & \textbf{0.703} $\pm$ 0.0040 & \textbf{0.696} $\pm$ 0.0029 \\
                    PA  & 0.672 $\pm$ 0.0017 & \textbf{0.695} $\pm$ 0.0023 & 0.688 $\pm$ 0.0022 \\
                    TX  & 0.690 $\pm$ 0.0029 & \textbf{0.712} $\pm$ 0.0028 & \textbf{0.712} $\pm$ 0.0027 \\
                    \hline
                    avg & 0.688 & \textbf{0.698}  & 0.692 \\
                    \hline
    \end{tabular}}
    \label{tab:test_real_results}
\end{table}

\noindent\textbf{Results} Table~\ref{tab:test_real_results} reports the mean and standard deviation of accuracies for 10 different test splits. For the F1 scores of the same experiment, see Appendix F.  Using all features leads to the best in-domain performance (see Appendix F), but not necessarily the best out-of-domain performance. Dropping the ambiguous features hurts predictive power in limited feature models, but helps with robustness varies across the states: these limited models even perform better on 2021 data. 
Our proposed feature engineering using CIS achieves the best of both worlds, with the best mean out-of-domain accuracy of 0.698. It also achieves close to the best out-of-domain accuracy for 8 out of 9 states. 

\section{Discussion}\label{sec:conclusion}
In this paper we studied the challenging problem of building models that are robust to distribution shift when causes and effects of the target variable are unmeasured. Among the observed noisy proxies, we showed how to perform feature selection based on conditional independence tests and knowledge about some seed nodes. 

After bootstrapping, we often have a significant number of ambiguous proxies, which have components that are both helpful and hurtful to our model's robustness. Through CIS, however, we showed how to isolate robust predictive power from these ambiguous proxies using auxiliary learning tasks. We proved that including these engineered features safely increases robustness in our setting, while also improving accuracy. In our experiments on real census data under shifts due to the pandemic, we showed that the engineered features provided benefits for most states over using the ambiguous features directly or completely ignoring them. While our theoretical framework is involved, these experiments demonstrate improvements outside of our assumptions.

\paragraph{Relaxation of Assumptions}
A number of our assumptions can be softened. One softening of systemic sparsity would involve allowing edges within $\vec{U}$ so long as their dependence is relatively weak. Such a relaxation would involve using mutual information (or correlation) thresholds instead of independence tests. Sparsity assumptions may also be relaxed by building on ideas from mixtures of DAG structures like ~\citep{gordon2021identifying}.

The strongest assumption is that of separable ambiguous proxies. Under a softening of the separability assumption, we cannot guarantee that we have isolated only robust information from our ambiguous proxy -- some unstable information associated with $\Ubad$ may slip through. However, degrees of separability may still guarantee the benefit of the engineered feature.

While separability corresponds to invertability with linear functions, there are many examples of nonlinear that are separable. For example, when the effects of two causes have significantly different magnitudes they can be easily disentangled, such as fine and hyper-fine structures in atomic energy levels. Work on data fission \citep{leiner2022data} may provide valuable insights to help understand the degrees of separability for different choices of functions.

\clearpage
\bibliography{biblio}
\clearpage

\appendix
\section{Additional SCM Background}
\label{apx:d-separation background}
For any DAG $\G = (\vec{V}, \vec{E})$, we call $\vec{P} \subseteq \vec{E}$ a \textbf{path} if it connects $A$ and $B$ with no repeated vertices. The path is \textbf{directed} if it obeys the directions of the edges and \textbf{undirected} if it does not.
Both directed and undirected paths in a causal DAG can result in dependencies between variables. To understand the conditions for dependence/independence ($d$-connection/$d$-separation) \cite{pearl1988probabilistic}, we will use the concepts of \textbf{active} and \textbf{inactive} paths, which are defined relative to a conditioning set \cite{pearl2009causality, peters2017elements}. Intuitively, whether a path is active or not indicates whether it ``carries dependence'' between the variables.

For an empty conditioning set, a path between $A$ to $B$ is active if it is directed or if it is made up of two directed paths from a common cause along that path. In the same unconditioned setting, \textbf{inactive paths} are paths that contain a \textbf{collider}, i.e. a vertex for which the path has two inward pointing arrows.

When we are given a conditioning set $\vec{Z}$, conditional dependencies differ from unconditional ones. Active paths can be \textbf{blocked} (thus becoming inactive paths) if some vertex $Z$ along the path between $A$ to $B$ is included in $\vec{C}$. Similarly, inactive paths with a collider $C$ can become \textbf{unblocked} by including $C$ or some descendant of the collider variable in the conditioning set $\vec{Z}$.
If two variables $A, B$ contain no active paths (they may contain inactive paths), then we say they are $d$-separated ($A \indep_d B \given \vec{Z}$). If two variables contain at least one active path for a conditioning set $\vec{Z}$, we say that they are $d$-connected.

\cite{pearl1988probabilistic} uses structural causal models to justify the \emph{local Markov condition}, which means that $d$-Separation always implies independence and allows DAG structures to be factorized. It is possible that two $d$-connected variables by chance exhibit some unexpected \emph{statistical} independence. The assumption of faithfulness \cite{spirtes2000causation} ensures that $d$-connectedness implies statistical dependence. This assumptions is popular in the causal discovery literature, but we will not need it for our theoretical results.
\section{Information Theory Preliminaries}
\label{apx: info theory prelim}
\begin{lemma}[Chain Rule, \citep{cover1999elements}]\label{lem:chainrule}
For sets of variables $\vec{A}, \vec{B}$, and subset $\vec{B}' \subset \vec{B}$
\begin{equation}
    \I(\vec{A} : \vec{B}) = \I(\vec{A}: \vec{B}') + \I(\vec{A}: \vec{B} \setminus \vec{B'} \given \vec{B}')
\end{equation}
\end{lemma}
\begin{definition}[\citep{cover1999elements}]
For sets of variables $\vec{A}, \vec{B}, \vec{C}$, the \textbf{interaction information} is defined,
\begin{equation}
    \I(\vec{A}: \vec{B}: \vec{C}) := \I(\vec{A} : \vec{B}) - \I(\vec{A} : \vec{B} \given \vec{C}).
\end{equation}
\end{definition}
A key property of interaction information is that it is symmetric to permutations in its three inputs,
\begin{equation}
        \I(\vec{A}: \vec{B}: \vec{C}) = \H(\vec{A}, \vec{B}, \vec{C}) + \H(\vec{A}) + \H(\vec{B}) + \H(\vec{C}) - \H(\vec{A}, \vec{B}) - \H(\vec{B}, \vec{C}) - \H(\vec{C}, \vec{A}).
\end{equation}

Another key property is that interaction information can be either positive or negative, differing from mutual information which is non-negative. The following lemmas will describe two common situations in which we can expect positive and negative interaction information.

\begin{lemma}\label{lem:positiveII}
    Given three sets of random variables $\vec{A}, \vec{B}, \vec{C}$ if $\vec{A} \indep \vec{C} \given \vec{B}$ then $\I(\vec{A}:\vec{B}:\vec{C}) \geq 0$.
\end{lemma}
Graphically, Lemma~\ref{lem:positiveII} represents a situation where conditioning on $\vec{B}$ $d$-separates $\vec{A}$ and $\vec{C}$ (i.e. $\vec{B}$ is a \emph{separating set} of $\vec{A}$ and $\vec{B}$). Hence, a sufficient condition to utilize Lemma~\ref{lem:positiveII} is $\vec{A} \doubleactivepathNB \vec{B} \doubleactivepathBN \vec{C}$, in which case the inequality becomes strict. The symmetry of interaction information means that it is not important which set of variables is the separating set. Conveniently $\I(A: B: C) \geq 0 \Rightarrow \I(A:B \given C) \leq \I(A:B)$, meaning we can ``drop'' conditioned variables in our upper bounds.

The \textbf{data processing inequality} uses each ``step'' of an active path to upper bound the mutual information.
\begin{lemma}[Data Processing Inequality (modified from \cite{cover1999elements})]
If $\vec{A} \indep \vec{C} \given \vec{B}, \vec{D}$ then
\begin{align}
       \I(\vec{A} : \vec{C} \given  \vec{D}) & \leq \min(\I(\vec{A}:\vec{B} \given \vec{D}), \I(\vec{B}:\vec{C} \given \vec{D})) \nonumber \\
       &\leq \H(\vec{B} \given \vec{D}).
\end{align}
\label{lem: DPI}
\end{lemma}
\section{Redundancy Bounds without Functional Assumptions} \label{apx:redundancy bounds without func assumtions}
Without the functional assumptions of our framework, we can only provide upper bounds on the context sensitivity.

To decompose the mutual information between sets of vertices $\vec{M}$ we will use the chain rule:
\begin{equation}
    \I(Y:\vec{M} \given \vec{X}) = \I(Y:M \given \vec{X}) + \I(Y:\vec{M} \setminus \{M\} \given \vec{X}, M)
\end{equation}
Repeated application of the chain rule accumulates conditioned $M$ variables for an arbitrary ordering. This conditioning represents the interactive nature of distribution shift mechanisms. We will now consider the context sensitivity of an arbitrary context variable conditioned on some arbitrary set of previously considered $\vec{M}' \subset \vec{M}$. We will be able to drop this conditioning on $\vec{M}'$ in our context sensitivity bounds.

\begin{lemma}
\label{lem: applied DPI}
For some $U_i \in \vec{U}$, and $\vec{M}' \subseteq \vec{M}$, if $M_i \doubleactivepathNB U_i \doubleactivepathBN Y$, then
\begin{equation*}
    \I(M_i:Y \given \vec{X}, \vec{M}') \leq \H(U_i \given \vec{X}) = \H(U_i) - \I(U_i:\vec{X}).
\end{equation*}
\end{lemma}
\begin{proof}
Using the data processing inequality (Lemma~\ref{lem: DPI}),
\begin{align*}
    \I(M_i:&Y \given \vec{X}, \vec{M}') \\
    &\leq \min (\I(U_i:M_i \given \vec{X}, \vec{M}'), \I(U_i:Y \given \vec{X}, \vec{M}'))\\
    &\leq H(U_i \given \vec{X}, \vec{M}') \\
    &\leq H(U_i \given \vec{X})\tag*{\qedhere}
\end{align*}
\end{proof}

Lemma~\ref{lem: applied DPI} gives an information-theoretic quantification of the notion of $d$-separation, showing that reducing context sensitivity involves selecting $\vec{X}$ with high $\Ugood$ redundancy. An invariant set is an extreme case of this rule, in which $\vec{X}$ has full redundancy with $\Ugood$.

\begin{lemma}
\label{lem: Collider DPI}
For $U_i \in \vec{U}, \vec{X} \subseteq \vec{V}$ let $\vec{X}' = \vec{X} \cap \Ch(U_i)$. If $U \in \Ubad$ then 
\begin{equation*}
    \I(M_i:Y \given \vec{X}, \vec{M}') \leq \I(U_i:\vec{X}' \given Y) \leq \I(U_i:\vec{X}').
\end{equation*}
\end{lemma}
Lemma~\ref{lem: Collider DPI} quantitatively states that we should avoid redundancy with $\Ubad$.
\begin{proof}
$M_i \indep_d Y$ by the conditions of the lemma. Because $\vec{X} \setminus \vec{X}'$ has no vertices in $\Ch(U_i)$ and there are no other descendants, $M_i \indep_d Y \given (\vec{X} \setminus \vec{X}')$. 
\begin{equation}
\begin{split}
    \I(M_i:Y \given \vec{X}, \vec{M}') &= -\I(M_i:Y:\vec{X}' \given  \vec{M}')\\
        &=\I(M_i: \vec{X}' \given Y, \vec{M}') \\
        &- \I(M_i:\vec{X} \given  \vec{M}')\\
    &\leq \I(M_i: \vec{X}' \given Y, \vec{M}').
\end{split}
\end{equation}

$M_i \doubleactivepathNB U_i \doubleactivepathBN \vec{X}'$ because $U_i$ is the only descendant of $M_i$. The DPI gives,
\begin{equation}
    \I(M_U: \vec{X}' \given Y,  \vec{M}') \leq \I(U: \vec{X}' \given Y,  \vec{M}').
\end{equation}
We apply Lemma~\ref{lem:positiveII} because $\vec{X}' \doubleactivepathNB U_i \doubleactivepathBN \vec{M}'$
\begin{equation}
    \I(M_U: \vec{X}' \given Y,  \vec{M}') \leq \I(U: \vec{X}' \given Y).
\end{equation}
We also have $\vec{X}' \doubleactivepathNB U_i \doubleactivepathBN Y$ from structural sparsity, so applying Lemma~\ref{lem:positiveII} gives the final loosest bound.
\end{proof}
\section{Guaranteeing Redundancy} \label{apx: guarnateeing redundancy}
Though we cannot ever measure the relevance of a proxy with unobserved $\vec{U}$, Lemma~\ref{lemma:mutualinf} shows that we can harness the graphical structure to obtain lower bounds.
\begin{lemma}[Unobserved common-cause information]\label{lemma:mutualinf}
Given a causal DAG $\mathcal{G} = (\vec{V}, \vec{E})$, for any $U, V_i, V_j \in \vec{V}$ that satisfy $V_i \doubleactivepathNB U \doubleactivepathBN V_j$, then  $\I(V_i, V_j: U) \geq \I(V_i:V_j)$.
\end{lemma}
\begin{proof}
A visualization of this proof is given in Figure~\ref{fig:MI_bound_proof}. Colors are added to the equations in the proof to match this figure. Begin with the definition of mutual information.
\begin{equation}
    \I(V_i, V_j:U) = \H(V_i, V_j) - \H(V_i, V_j \given U)
\end{equation}
We can expand the joint entropy of both terms as follows,
\begin{align}
    \H(V_i, V_j) =& \H(V_i \given V_j) + \H(V_j \given V_i) + \I(V_i:V_j) \label{eq:expand_ABC}\\
    \H(V_i, V_j \given U) =& \H(V_i \given V_j, U) + \H(V_j \given V_i, U) + \underbrace{\I(V_i:V_j \given U)}_{=0 \text{ because } V_i \indep_d^{\mathcal{G}} V_j \given U} \label{eq:expand_ABDC}
\end{align}
Together, Equations \ref{eq:expand_ABC} and \ref{eq:expand_ABDC} give:
\begin{align*}
    \begin{split}
    \I(V_i, V_j:U) =& \color{red}{\H(V_i \given V_j, U)} \color{black}{ + } \color{blue}{\H(V_j \given V_i)} \color{black}{ + } \color{purple}{\I(V_i: V_j)} \color{black}{ - } \color{red}{\H(V_i \given U, V_j)} \color{black}{ - } \color{blue}{\H(V_j \given U, V_i)}
    \end{split}\\
    =& \color{red}{\I(V_i:U \given V_j)} \color{black}{ + } \color{blue}{\I(V_j:U \given V_i)} \color{black}{ + } \color{purple}{\I(V_i: V_j)}\\
    \geq& \color{purple}{\I(V_i: V_j)}
\end{align*}
\end{proof}
\def\firstcircle{(0,0.5) circle (1.7cm)}
\def\secondcircle{(1.9, -1) circle (1.5cm and 3.0cm)}
\def\thirdcircle{(-1.9, -1) circle (1.5cm and 3.0cm)}
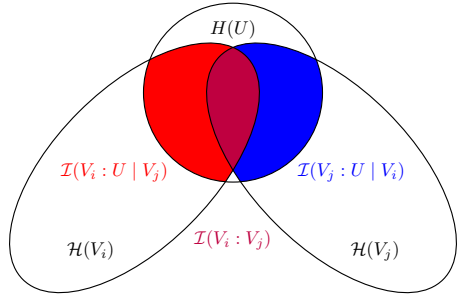
\begin{figure}[h]
\setlength{\belowcaptionskip}{-10pt}
\centering
    \centering
    \scalebox{.7}{
    \begin {tikzpicture}[-latex ,auto ,node distance =1.5 cm and 1.5 cm ,on grid , semithick, state/.style ={ circle, draw, minimum width =.5 cm}]
    \begin{scope}
      \clip \firstcircle;
      \fill[rotate around= {45:(1.8, -1)}, blue] \secondcircle;
    \end{scope}
    
    \begin{scope}
      \clip \firstcircle;
      \fill[rotate around= {-45:(-1.8, -1)}, red] \thirdcircle;
    \end{scope}
    
    \begin{scope}
      \clip[rotate around= {45:(1.8, -1)}] \secondcircle;
      \fill[rotate around= {-45:(-1.8, -1)}, purple] \thirdcircle;
    \end{scope}
    \draw \firstcircle node[above=.9] {$H(U)$};
    \draw[rotate around={45:(1.8, -1)}] \secondcircle;
    \node (hac) at (-2.7, -2.5) {$\H(V_i)$};
    \draw[rotate around={-45:(-1.8, -1)}] \thirdcircle; 
    \node (hbc) at (2.7, -2.5) {$\H(V_j )$};
    \node[blue] (hij) at (2.25, -1) {$\I(V_j:U \given V_i)$};
    \node[red] (hji) at (-2.25, -1) {$\I(V_i:U \given V_j)$};
    \node[purple] (i) at (0, -2.3) {$\I(V_i: V_j )$};
    
        \end{tikzpicture}
        }
    \caption{A visual proof for Lemma~\ref{lemma:mutualinf}.}
    \label{fig:MI_bound_proof}
\end{figure}

Hence, we can lower bound $\Fisolate{V_G}{U_A}$'s hold on good information in $U_A^{(G)}$ using Lemma~\ref{lemma:mutualinf},
\begin{equation}
   \I((V_i, \Fisolate{V_i}{V_A}):\Pa(V_i)) \geq \I(V_A:V_i). 
\end{equation}
\section{Deferred Proofs}\label{apx: deferred proofs}

\subsection{Proof of Lemma~\ref{lem: d conn but not dependent}}
\begin{proof}
 We can rewrite the conditional mutual information making use of $U_1 \indep U_2$ as follows.
\begin{equation*}
\begin{aligned}
        \I(U_1:U_2 \given V) &= -\I(U_1 : U_2 : V)\\
        &=-\I(U_1:V) + \I(U_1:V \given U_2)\\
        &=-\I(U_1:V^{(U_1)}) + \I(U_1:V^{(U_2)} \given U_2)\\
        &=-\I(U_1:V^{(U_1)}) + \I(U_1:V^{(U_1)})= 0.
\end{aligned}\qedhere
\end{equation*}
\end{proof}

\subsection{Proof of Lemma~\ref{lem: setting redundancy}}
\begin{proof}
$\I(U:\vec{X}) = \H(U) - \H(U \given \vec{X})$. If at least one child of $U$ is conditioned on and transmits (i.e. in $V \in \Ch(U) \cap \vec{X}$ and $v_{\vec{x}} \neq \phi)$, then $\H(U \given \vec{x}) = 0$. Otherwise, $\H(U \given \vec{X}) = \H(U)$ because all $X \in \vec{X} \setminus \Ch(U)$ with active paths to $U$ must go through colliders in $\Ch(U)$ - all of which are not in $\vec{X}$ or not transmitting.
\end{proof}

\subsection{Proof of Lemma~\ref{lem: new applied DPI}}
\begin{proof}
 $\I(M_i:Y \given \vec{X}) = \sum_{\vec{X}} \I(M_i:Y \given \vec{X})$ is zero unless $(M_i, U_i)$ and $(U_i, Y)$ both transmit. Furthermore, the DPI gives $\I(M_i:Y \given \vec{X}) \leq \H(U_i \given \vec{X})$, which is zero if any of the edges from $U_i \rightarrow X$ for $X \in \vec{X}$ transmit.
\end{proof}

\subsection{Proof of Lemma~\ref{lem: new Collider DPI}}
\begin{proof}
 $\I(M_i:Y \given \vec{X}) = \sum_{\vec{X}} \I(M_i:Y \given \vec{X})$ is zero unless both $(M_i, U_i)$ and $(U_i, Y)$ transmit and at least one of $(U_i, X)$ transmits for $X \in \vec{X}$, in which case $\I(M_i:Y \given \vec{x}) = \I(M_i:Y \given U_i)$.
\end{proof}

\subsection{Proof of Lemma~\ref{lem:conditioning on y separates sets}}
\begin{proof}
($\Rightarrow$) We will prove this with the contrapositive. If there is no shared parent between $V_i$ and $V_j$, then all active paths must go through $Y$. However, because at least one of $V_i, V_j$ is not connected to a cause, all paths between $V_i$ and $V_j$ cannot have a collider at $Y$ (through two causes). This means conditioning on $Y$ blocks the remaining paths.

($\Leftarrow$) If there is a shared parent $U$ between $V_i$ and $V_j$, then $V_i \leftarrow U \rightarrow V_j$ is an active path, $d$-connecting the two vertices. If $V_i$ and $V_j$ each have corresponding parents $U_i, U_j \in \Pa(Y)$, then $V_i \leftarrow U_i \rightarrow Y \leftarrow U_j \rightarrow V_j$ is an active path conditioned on $Y$.
\end{proof}

\subsection{Proof of Lemma~\ref{lem: bootstraping}}
\begin{proof}
 The proof follows from Lemma~\ref{lem:conditioning on y separates sets}. Adjacent edges in $\G_Y$ either indicate shared parents or that both vertices have a (potentially different) cause of $Y$ as their parent. 
 
 If $V_i$ and $V_j$ share a parent, then $\Pa(V_i) \subseteq \Ugood$ implies $\Ugood \cap \Pa(V_j) \neq \emptyset$, so $V_j$ has at least one ``good'' parent. The symmetric argument holds for $\Pa(V_i) \subseteq \Ubad$.
 
 If $V_i$ and $V_j$ both have at least one causal parent, then we know $V_i, V_j \not \in \Vbad$. We know both vertices have at least one good $U$ as a parent, so it trivially follows that both are either in $\Vgood$ or $\Vambig$.
 \end{proof}
 
 \subsection{Proof of Theorem~\ref{thm: proxy bootstrapping works}}
\begin{proof}
The only potential conditioned collider is $Y$, which is not allowed to be separable by partial faithfulness. Hence, partial faithfulness guarantees that we can construct $\G_Y$ from conditional independence tests because $d$-connection implies dependence between $V$.

 The requirements on $\vec{V}^*$ given by the theorem ensure that every $V \in \vec{V}$ has at least one label from an adjacency to a $V^* \in \vec{V}^*$ in $\G_Y$.
 
 The algorithm adds ``good'' labels to all vertices with a known good parent and ``bad'' labels to all vertices with a known bad parent. Therefore, all ``ambigious'' vertices are correctly labeled. 
 
 We now only need to guarantee that that the ``good'' and ``bad'' vertices are not ambiguous. If $V$ were ambiguous, it would be connected to a $U$ of the opposite label (i.e. a ``good'' vertex would be connected to a bad $U$). Such a $U$ would have at least one $V^* \in \vec{V}^* \cap \Ch(U)$ which would be adjacent to $V$ and have given $V$ the label of $U$, a contradiction.
 \end{proof}
 
 \subsection{Proof of Lemma~\ref{lem: new applied DPI engineering}}
 \begin{proof}
$\I(M_i:Y \given \vec{X}) = \sum_{\vec{x} \in \vec{X}} \Pr(\vec{x}) \I(M_i:Y \given \vec{x})$. Now, we have \[\I(M_i:Y \given \vec{x}) = \H(M_i \given \vec{x}) = \H(U_i \given \vec{x})\] if both $(M_i, U_i)$ and $(U_i, Y_i)$ edges transmit, which occurs with probability $\alpha_{M_i, U_i}\alpha_{U_i, Y}$. Pulling this coefficient outside of the sum gives $\I(M_i:Y \given \vec{X}) = \alpha_{M_i, U_i}\alpha_{U_i, Y} \H(U_i \given \vec{x})$.

\end{proof}

\subsection{Proof of Lemma~\ref{lem: if we dont include information with bad u, then we are happy} }
\begin{proof}
    $U_i \in \Ubad$ means $M_i \doubleactivepathNC U_i\doubleactivepathCN Y$, so $\I(M_i: Y) = 0$.
    \begin{equation}
        \begin{aligned}
            \I(M_i:Y \given \vec{X}) &= -\I(M_i: Y: \vec{X})\\
            &\leq \I(M_i:\vec{X} \given Y)\\
            &\leq \I(U_i: \vec{X} \given Y) = 0
        \end{aligned}
    \end{equation}
    The final inequality comes from the data processing inequality.
\end{proof}

\subsection{Proof of Lemma~\ref{lem: isolation bound 2}}
\begin{proof}
Consider the function $F_C(V_A) = F_C(G, B)) = \Fisolate{V_G}{G}$. By definition,
\begin{align}
    \I(\Fisolate{V_G}{G}:V_G \given Y) &= \I(G:V_G \given Y)\\
    &= \I(V_A:V_G \given Y).
\end{align}
Hence, $F_C$ is in the feasible set of the optimization function defining isolation functions. Furthermore, $F_C(V_A)$ is only a function of $G$ and $G \indep U_B \given Y$, so we can also conclude that $F_C(V_A) \indep U_B$. This means
\begin{align*}
\H(F_C(V_A) \given Y) &=\H(F_C(V_A) \given U_B, Y)\\
 \leq& \H(\Fisolate{V_G}{V_A} \given U_B, Y)\\
 \leq& \H(\Fisolate{V_G}{V_A} \given Y) - \I(\Fisolate{V_G}{V_A} : U_B \given Y).
\end{align*}
Hence, if $\I(\Fisolate{V_G}{V_A} : U_B \given Y) > 0$, then $\H(F_C(V_A) \given Y) < \H(\Fisolate{V_G}{V_A} \given Y)$, contradicting the minimality of $\Fisolate{V_G}{V_A}$.
\end{proof}

\subsection{Proof of Theorem~\ref{thm: when help and do no harm}}
\begin{proof}
To shorten some equations, we will use \[\vec{F}(V_A):=\Fisolate{V_G}{V_A \given Y}.\]
We first show that Equation~\ref{eq: thm2} is sufficient for an improvement in relevance. We can expand the relevance of $\vec{X}^+$ as follows
\begin{equation}
\begin{split}
    \I(Y:\vec{X}^+) =& \I(Y:\vec{F}(V_A)) + \I(Y: V_G\given \vec{F}(V_A))\\
& \geq \I(Y: V_G\given \vec{F}(V_A))\\
&\geq \I(Y: V_G) - \I(Y: V_G : \vec{F}(V_A)).
\end{split}
\end{equation}
So, for guaranteed improvement in relevance ($\I(Y: \vec{X}^+) > \I(Y:V_G)$), we need negative $\I(Y: V_G : \vec{F}(V_A)) < 0$. Expanding,
\begin{equation}
    \I(Y: V_G : \vec{F}(V_A)) = \I(\vec{F}(V_A):V_G) - \I(\vec{F}(V_A):V_G \given Y).
\end{equation}
Thus, Equation~\ref{eq: thm2} gives us the exact condition needed for negative interaction information, guaranteeing improvement.

We can show that the context sensitivity is no worse by separately considering the context sensitivity with $\Pa(\Ubad)$ and $\Pa(\Ugood)$. We begin with $\Ubad$. Applying Lemma~\ref{lem: isolation bound 2},
\begin{equation}
    \I((V_G, \vec{F}(V_A)): \Ubad \given Y) = 0,
\end{equation}
which satisfies the conditions for Lemma~\ref{lem: if we dont include information with bad u, then we are happy} to ensure us that $\I(\Pa(\Ubad):Y \given \vec{X}^+) = 0$.

Now, consider an arbitrary for $M_G = \Pa(U_G) \in \Pa(\Ugood)$. Lemma~\ref{lem: new applied DPI engineering} tells us that
\begin{equation*}
    \I(M_G:Y \given \vec{X}^+) = \alpha_{M_G:U_G} \alpha_{U_G, Y} H(U_G \given \widetilde{\Ch}_{\vec{X}^{+}}(U_G)).
\end{equation*}
We then observe that $H(U_G \given \widetilde{\Ch}_{\vec{X}^{+}}(U_G)) \leq H(U_G \given \vec{X})$ because entropy is submodular, which leads us to conclude,
\begin{equation*}
     \I(M_G:Y \given \vec{X}^+) \leq \I(M_G:Y \given \vec{X}).
\end{equation*}
This completes the proof.
\end{proof}

\section{Experiments}
\subsection{Synthetic experimental setup}\label{sec:appendix_synthethic}
 $M_G$ and $M_B$ are drawn from normal distributions with mean $0$ and variable standard deviations. All other vertices (other than $Y$) are the average of their parents plus additional Gaussian noise $N(0, .2)$. $T_A\in \R^2$ is generated by applying a rotation matrix to $(T_A^{(G)}, T_A^{B})^T$\footnote{Many rotations were tried in our experiments with identical results, so we display results from a $45$ degree rotation.}. $Y$ indicates whether its parents  sum to a positive number with a $5\%$ probability of flipping randomly.

\subsection{F1 scores for real world experiment}
We give the F1 scores for the experiment described in Section~\ref{sec:real_data} in Table~\ref{tab:performance_via_f1}.
\begin{table}[htbp]
\centering
\small
\caption{Comparison of out-of-domain (2021) performance on predicting high income via F1 scores.}
\label{tab:performance_via_f1}
\begin{tabular}{|c|l l l|}
\hline
State & All Features & Engineered Features & Limited Features \\
\hline
CA & \textbf{0.684} & \textbf{0.683} & 0.676 \\
FL & \textbf{0.459} & 0.388 & 0.388 \\
GA & 0.541 & \textbf{0.626} & \textbf{0.624} \\
IL & 0.563 & \textbf{0.630} & \textbf{0.628} \\
NY & \textbf{0.688} & \textbf{0.690} & 0.662 \\
NC & \textbf{0.475} & 0.410 & 0.410 \\
OH & 0.519 & \textbf{0.581} & \textbf{0.580} \\
PA & 0.531 & \textbf{0.608} & \textbf{0.606} \\
TX & 0.554 & \textbf{0.619} & \textbf{0.619} \\
\hline
avg & 0.557 & \textbf{0.582} & 0.577 \\
\hline
\end{tabular}
\end{table}

\subsection{Real world experiment in-domain performance}\label{apx:real_data}
Here we provide the results of the in-domain accuracy for the experiment described in Sec.~\ref{sec:real_data}.
Recall, that we use US Census data and consider distributions shifts across time as suggested by~\cite{ding2021retiring}. 
Table~\ref{tab:test_2019_real_results} shows the accuracy on 2019 data on a held-out dataset (separate from the training split). We repeated the experiment 10 times on different training/testing splits and report the mean and standard deviation of the accuracy for the largest states in the U.S. As expected, using all features has the most predictive power for in-domain tasks.

\begin{table}[ht!]
    \centering
    \small
    \caption{Comparison of in-domain (2019) performance on predicting high income via Accuracies.}
    \begin{tabular}{|c | l l l |  }
     \hline
     State & All Features & Engineered Features & Limited Features   \\ 
    \hline
CA  & \textbf{0.713} $\pm$ 0.0010 & \textbf{0.710} $\pm$ 0.0012 & 0.691 $\pm$ 0.0011 \\
FL  & \textbf{0.700} $\pm$ 0.0014 & 0.693 $\pm$ 0.0020 & 0.694 $\pm$ 0.0017 \\
GA  & \textbf{0.708} $\pm$ 0.0025 & \textbf{0.708} $\pm$ 0.0036 & \textbf{0.707} $\pm$ 0.0036 \\
IL  & \textbf{0.689} $\pm$ 0.0023 & \textbf{0.690} $\pm$ 0.0039 & \textbf{0.685} $\pm$ 0.0021 \\
NY  & \textbf{0.705} $\pm$ 0.0024 & 0.698 $\pm$ 0.0022 & 0.687 $\pm$ 0.0076 \\
NC  & \textbf{0.713} $\pm$ 0.0020 & 0.703 $\pm$ 0.0049 & 0.700 $\pm$ 0.0028 \\
OH  & \textbf{0.717} $\pm$ 0.0029 & \textbf{0.716} $\pm$ 0.0042 & \textbf{0.712} $\pm$ 0.0033 \\
PA  & \textbf{0.702} $\pm$ 0.0028 & \textbf{0.701} $\pm$ 0.0027 & 0.695 $\pm$ 0.0026 \\
TX  & \textbf{0.708} $\pm$ 0.0019 & \textbf{0.705} $\pm$ 0.0025 & \textbf{0.706} $\pm$ 0.0022 \\
\hline
avg & \textbf{0.706} &  0.703  &  0.697 \\
\hline
    \end{tabular}
    \label{tab:test_2019_real_results}
\end{table}

\end{document}